\documentclass[lettersize,journal]{IEEEtran}
\usepackage{amsmath,amsfonts}
\usepackage{algorithmic}
\usepackage{algorithm}
\usepackage{array}
\usepackage[caption=false,font=normalsize,labelfont=sf,textfont=sf]{subfig}
\usepackage{textcomp}
\usepackage{stfloats}
\usepackage{url}
\usepackage{color}
\usepackage{multirow}
\usepackage{tabularray}
\usepackage{verbatim}
\usepackage{graphicx}
\usepackage[utf8]{inputenc}
\RequirePackage[colorlinks]{hyperref}

\usepackage{bbding}
\usepackage{booktabs}
\usepackage{cite}
\begin{document}

\title{Camouflaged Object Tracking: A Benchmark}

\author{ Xiaoyu Guo, Pengzhi Zhong, Hao Zhang, Defeng Huang, Huikai Shao, Qijun Zhao, and Shuiwang Li$^*$\thanks{*Corresponding author.}

\thanks{Xiaoyu Guo, Pengzhi Zhong, Hao Zhang, Defeng Huang, and Shuiwang Li are with the College of Computer Science and Engineering, Guilin University of Technology, China, 541006 and Guangxi Key Laboratory of Embedded Technology and Intelligent System, Guilin University of Technology, China, 541004 (e-mail: guoxiaoyu@glut.edu.cn; zhongpengzhi@glut.edu.cn; zhanghao@glut.edu.cn;  huangdefeng@glut.edu.cn; 
lishuiwang0721@glut.edu.cn).}
\thanks{Huikai Shao is with the School of Automation Science and Engineering, Xi'an Jiaotong University, Xi'an, Shaanxi 710049, China (e-mail: shaohuikai@xjtu.edu.cn).}
\thanks{Qijun Zhao is with the College of Computer Science, Sichuan University, Sichuan 610065, China (e-mail: qjzhao@scu.edu.cn).}

}

\markboth{}%
{}

\maketitle

\begin{abstract}

Visual tracking has seen remarkable advancements, largely driven by the availability of large-scale training datasets that have enabled the development of highly accurate and robust algorithms. While significant progress has been made in tracking general objects, research on more challenging scenarios, such as tracking camouflaged objects, remains limited. Camouflaged objects, which blend seamlessly with their surroundings or other objects, present unique challenges for detection and tracking in complex environments. This challenge is particularly critical in applications such as military, security, agriculture, and marine monitoring, where precise tracking of camouflaged objects is essential. To address this gap, we introduce the Camouflaged Object Tracking Dataset (COTD), a specialized benchmark designed specifically for evaluating camouflaged object tracking methods. The COTD dataset comprises 200 sequences and approximately 80,000 frames, each annotated with detailed bounding boxes. Our evaluation of 20 existing tracking algorithms reveals significant deficiencies in their performance with camouflaged objects. To address these issues, we propose a novel tracking framework, HiPTrack-MLS, which demonstrates promising results in improving tracking performance for camouflaged objects. COTD and code are avialable at  \url{https://github.com/openat25/HIPTrack-MLS}.
\end{abstract}

\begin{IEEEkeywords}
Visual tracking; Camouflaged object tracking; Benchmark
\end{IEEEkeywords}

\section{Introduction}

\IEEEPARstart{I}{n}  recent years, the field of visual tracking has made significant advances, largely due to the widespread use of large-scale training datasets. The richness of these datasets has facilitated the development of high-precision and robust tracking algorithms \cite{danelljan2020probabilistic,gao2022aiatrack,lijia2020adaptive,marvasti2021deep,sm2021review,yuan2023active,zhang2020correlation,zhu2022fast}. Although substantial progress has been made in general object tracking, research into more specialized and challenging scenarios remains limited. Among these, tracking camouflaged objects is particularly difficult because these targets seamlessly blend into their surroundings, making them hard to detect with the naked eye and significantly increasing tracking complexity \cite{cuthill2019camouflage,xiao2016distractor}.

\begin{figure}[t]
	\centering
	{
		\begin{minipage}[t]{0.475\textwidth}
			\centering
			\includegraphics[width=1\textwidth,height=0.2\textwidth]{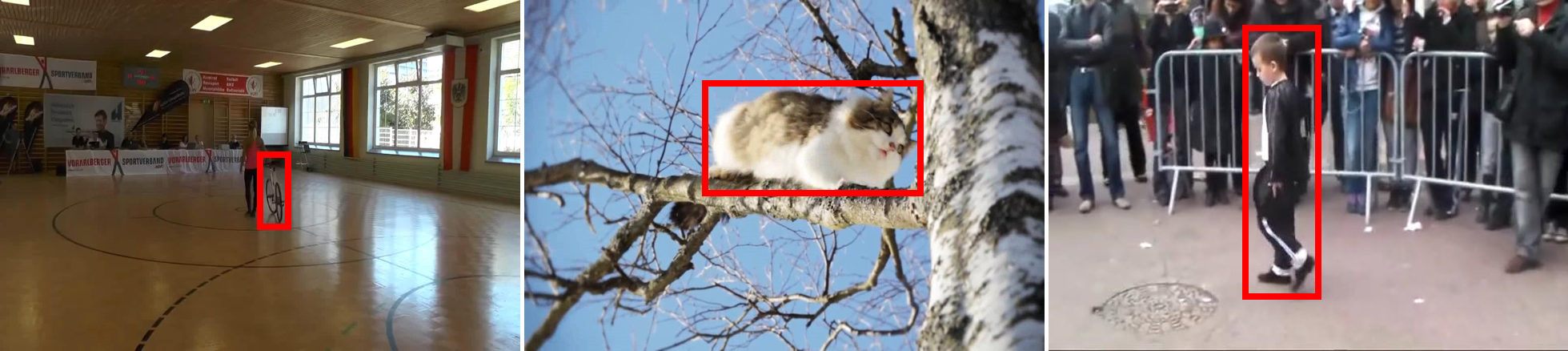}
			\centerline{\footnotesize (a) Example of generic object tracking.}
	\end{minipage}}
	{
		\begin{minipage}[t]{0.475\textwidth}
			\centering
			\includegraphics[width=1\textwidth,height=0.22\textwidth]{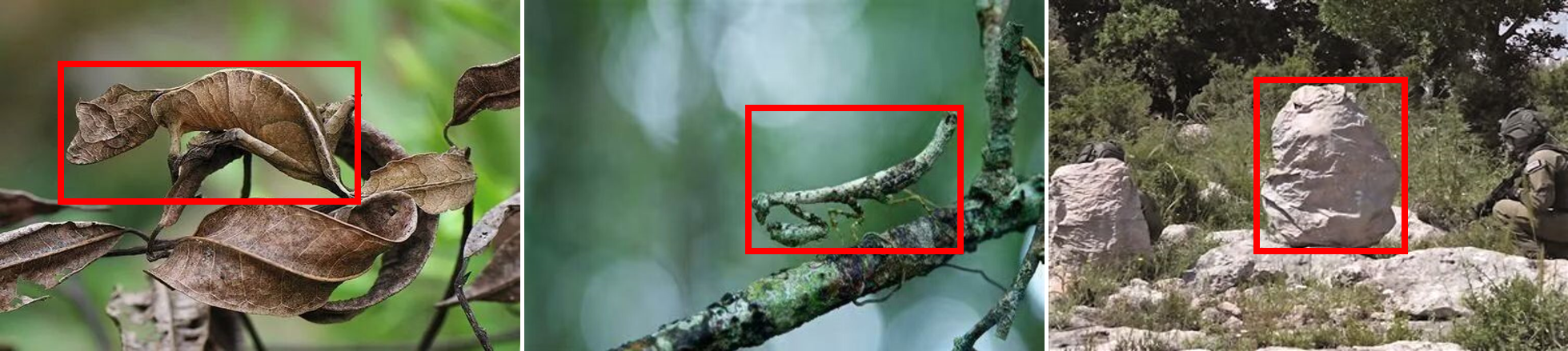}
			\centerline{\footnotesize (b) Example of camouflaged object tracking.}
	\end{minipage}}
	\caption{Generic object tracking (a) and camouflaged object tracking (b). Compared to generic object tracking, tracking camouflaged objects is more challenging due to their tendency to blend into the background, resulting in indistinguishable visual features. }
	\label{fig:exa}\vspace{-0.25in}
\end{figure}

Camouflaged objects play a crucial role in several critical applications across various domains. For instance, in the military and defense sector, the ability to identify and track camouflaged objects, such as hidden weapons or concealed enemy units, is vital for battlefield awareness and operational success \cite{hogan2016dazzle,liu2019concealed}. In the field of autonomous driving, camouflaged objects, such as animals or debris that blend into the road environment, pose significant challenges for vehicle perception systems \cite{zheng2024physical,ranjan2019competitive,cheng2019noise}. 
In medical imaging, camouflaged objects often take the form of pathological regions, such as tumors that are difficult to distinguish from surrounding healthy tissue \cite{fan2020pranet,wu2021jcs}. Additionally, in wildlife monitoring and ecological research, camouflaged animals often need to be tracked without disturbing their natural behavior \cite{cai2023semi,pembury2020camouflage,lee2023artificial}.

Distinguishing camouflaged objects from backgrounds is crucial but challenging due to factors like background clutter, boundary blurring, texture mimicry, color similarity, varying lighting, and occlusion \cite{mei2021camouflaged,cuthill2019camouflage,xiao2016distractor}. Addressing these challenges can significantly advance applications involving camouflaged objects \cite{cheng2022implicit,liu2023aerial,liu2023unsupervised,mondal2017partially,lv2021simultaneously}. Recent computer vision techniques, particularly in Camouflaged Object Detection (COD) and Camouflaged Object Segmentation (COS), have made notable progress. COD identifies objects blending into their surroundings, while COS further separates them from backgrounds by classifying each pixel \cite{fan2020camouflaged,liu2021camouflaged,yan2021mirrornet}. Ongoing research in these areas continues to push boundaries, promising innovations to tackle real-world challenges related to camouflaged objects.

Deep learning has significantly advanced COD and COS, with methods like MirrorNet \cite{yan2021mirrornet}, PFNet \cite{mei2021camouflaged}, and S-MGL \cite{zhai2022mgl} improving accuracy by learning from large datasets. Benchmark datasets such as NC4K \cite{lv2021simultaneously} and COD10K \cite{fan2020camouflaged} have been crucial for this progress.
However, Camouflaged Object Tracking (COT), which involves consistent identification of camouflaged objects in videos, remains under-explored. COT requires refined feature extraction and sophisticated algorithms to handle changes in position, scale, and appearance, as well as to manage uncertainties and coexisting objects \cite{wang2021deep, wang2020displacement, zhong2019unsupervised}. 
A critical gap exists: there is no specialized dataset for COT. This absence limits the development and evaluation of COT algorithms, as existing datasets may not capture the unique challenges of camouflaged objects. Without a dedicated dataset, researchers struggle to test their models against these specific nuances.

In this work, we take the first step in addressing the challenging task of tracking camouflaged objects by introducing the Camouflaged Object Tracking Dataset (COTD). COTD features a diverse range of camouflaged objects across various environments, specifically designed to push the boundaries of camouflaged object tracking algorithms. The dataset includes 200 sequences totaling around 80,000 frames, with each sequence meticulously annotated with detailed bounding boxes and relevant attributes for comprehensive performance evaluation and analysis.
Fig. \ref{fig:exa} demonstrates the differences between generic object tracking and camouflaged object tracking. Compared with generic object tracking, where targets are more distinguishable from their backgrounds, tracking camouflaged objects is notably more challenging as these objects blend seamlessly with their surroundings. This blending creates ambiguous visual cues that significantly complicate the tracking process. Our extensive dataset captures a broad spectrum of real-world scenarios involving camouflaged objects, laying a solid foundation for developing and evaluating new tracking algorithms that adapt to these conditions. We also conducted extensive evaluations of 20 state-of-the-art tracking algorithms and found that current methods often struggle with the complexities introduced by camouflaged objects, indicating the need for substantial improvements. By releasing COTD, we aim to advance tracking technology to address the unique challenges posed by camouflaged targets and ultimately enhance the accuracy and reliability of visual tracking systems across various applications. The main contributions of this paper are as follows:

\begin{itemize}
\item We introduce COTD, the first benchmark for tracking camouflaged objects, with 200 meticulously annotated sequences. COTD presents unique challenges not addressed by existing benchmarks and may inspire further research in this area.

\item We thoroughly evaluate 20 state-of-the-art trackers on COTD, highlighting their limitations, setting performance benchmarks, and inspiring future research in tracking camouflaged objects.

\item We introduce HIPTrack-MLS, a novel tracker that leverages multi-level features to enhance performance, establishing a stronger baseline for future research in tracking camouflaged objects.

\end{itemize}

\section{Related Works}

\begin{figure*}[t]
  \centering
  \includegraphics[width=1\linewidth]{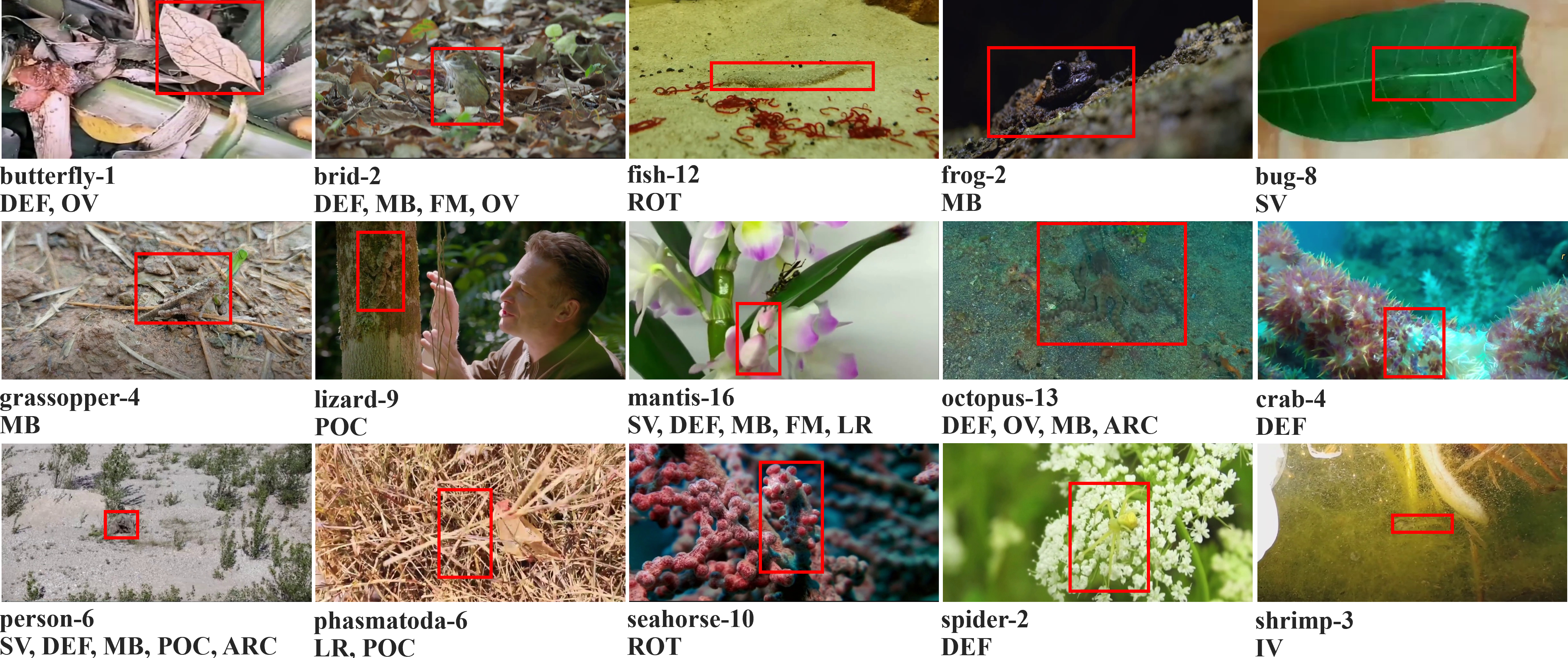}
  \caption{Some examples of annotations for COTD. Each sequence is annotated with
axis-aligned bounding boxes and attributes.}
  \label{fig:Anno}
\end{figure*}

\subsection {Visual Tracking Algorithms}
DCF-based trackers, like the early MOSSE filter \cite{bolme2010visual}, treat visual tracking as an online regression problem. Enhancements such as scale filters and regularization techniques have been introduced to improve robustness \cite{danelljan2016discriminative,li2015scale,danelljan2015learning,li2018learning,li2020asymmetric}. However, these trackers still struggle with complex conditions due to their reliance on handcrafted features.
To overcome these challenges, deep learning-based trackers have become popular. A key innovation was SiamFC \cite{bertinetto2016fully}, which used Siamese networks to learn a similarity function for generating response maps. Subsequent methods like SiamRPN++ \cite{li2019siamrpn++}, SiamR-CNN \cite{sio2020s2siamfc}, SiamCAR \cite{guo2020siamcar}, and Ocean \cite{zhang2020ocean} further enhanced accuracy and robustness.
Recently, Vision Transformers (ViTs) \cite{dosovitskiy2020image} have introduced new possibilities for visual tracking, offering potential for streamlining and unifying tracking frameworks \cite{dosovitskiy2020image,liu2021swin,DropTrack,ye2022joint}. For example, OStrack \cite{ye2022joint} combines feature extraction with relational modeling for efficient single-stream tracking, and HIPTrack \cite{cai2024hiptrack} introduces a history prompting network for more precise guidance.
However, most current research focuses on generic object tracking, with less attention given to specialized scenarios like camouflaged object tracking. In this work, we introduce a dataset for tracking camouflaged objects and enhance HIPTrack with multi-level features to address this challenge.

\subsection{Visual Tracking Benchmarks}

Currently, there are two main types of benchmark datasets: general benchmarks and specific benchmarks \cite{fan2021transparent}.
1)
\textbf{General Benchmarks: }
Many large-scale benchmarks have been introduced for training deep learning-based trackers. 
For example, TrackingNet \cite{muller2018trackingnet} aims to provide resources for training and evaluating general object tracking algorithms, containing over 30,000 sequences. GOT-10K \cite{huang2019got} includes over 10,000 videos from the WordNet semantic hierarchy to develop the generalization capability of trackers. LaSOT \cite{fan2019lasot} encompasses 1,400 long-term video sequences, with an additional 150 sequences and a new evaluation scheme.
2) \textbf{Specific Benchmarks: }
UAV123 \cite{benchmark2016benchmark} focuses on low-altitude drone object tracking, containing 123 sequences captured by drones. 
CDTB \cite{lukezic2019cdtb} and PTB \cite{song2013tracking} are designed to evaluate tracking performance in RGB-D videos, where D represents depth images. TOTB \cite{fan2021transparent} specializes in tracking transparent objects, collecting 225 videos from 15 categories of transparent objects. Recently, TSFMO \cite{zhang2022tracking} has been introduced for tracking small and fast-moving objects, while TRO \cite{Guo2024TrackingRO} focuses on tracking reflected objects.
Additionally, BihOT \cite{wang2024bihot} contains 49 videos and 42K frames, focusing on hyperspectral camouflage. UW-COT220 \cite{zhangunderwater} focuses on underwater camouflage.

\subsection{Dealing with camouflaged Objects in Vision}

Camouflaged objects blend into their surroundings due to similar textures, colors, and patterns, making them hard to distinguish. To address this, recent computer vision techniques have been developed. Fan et al. \cite{fan2020camouflaged} introduced Camouflaged Object Detection (COD) with the COD10K dataset and the SINet framework for efficient detection. Liu et al. \cite{liu2021camouflaged} proposed a scale-adaptive correlation filter for Camouflaged Object Segmentation (COS). Cheng et al. \cite{cheng2022implicit} developed a Video Camouflaged Object Detection (VCOD) framework with the MoCA-Mask dataset. Hu et al. \cite{hu2023visual} addressed visual camouflage in UAV surveillance. However, Camouflaged Object Tracking (COT), which involves consistent identification of camouflaged objects in videos, remains under-explored. To bridge this gap, we introduce the Camouflaged Object Tracking Dataset (COTD) to support algorithm development and evaluation for COT.

  \section{Construction of the COTD Dataset}
\subsection{Data Collection}

To thoroughly assess the performance of camouflaged object tracking algorithms, we develop a novel benchmark dataset known as the Camouflaged Object Tracking Dataset (COTD). COTD offers a diverse and high-quality platform for advancing and evaluating camouflaged object tracking technologies. Our rigorous and extensive data collection process involve extensive internet searches and scientific literature mining, initially gathering hundreds of potential high-quality videos. These videos encompass a variety of environments, including forests, grasslands, oceans, deserts, and snowfields, each carefully selected to exhibit distinct behaviors and motion patterns of camouflaged targets.

\begin{figure}[t]
  \centering
   \includegraphics[width=0.9\linewidth]{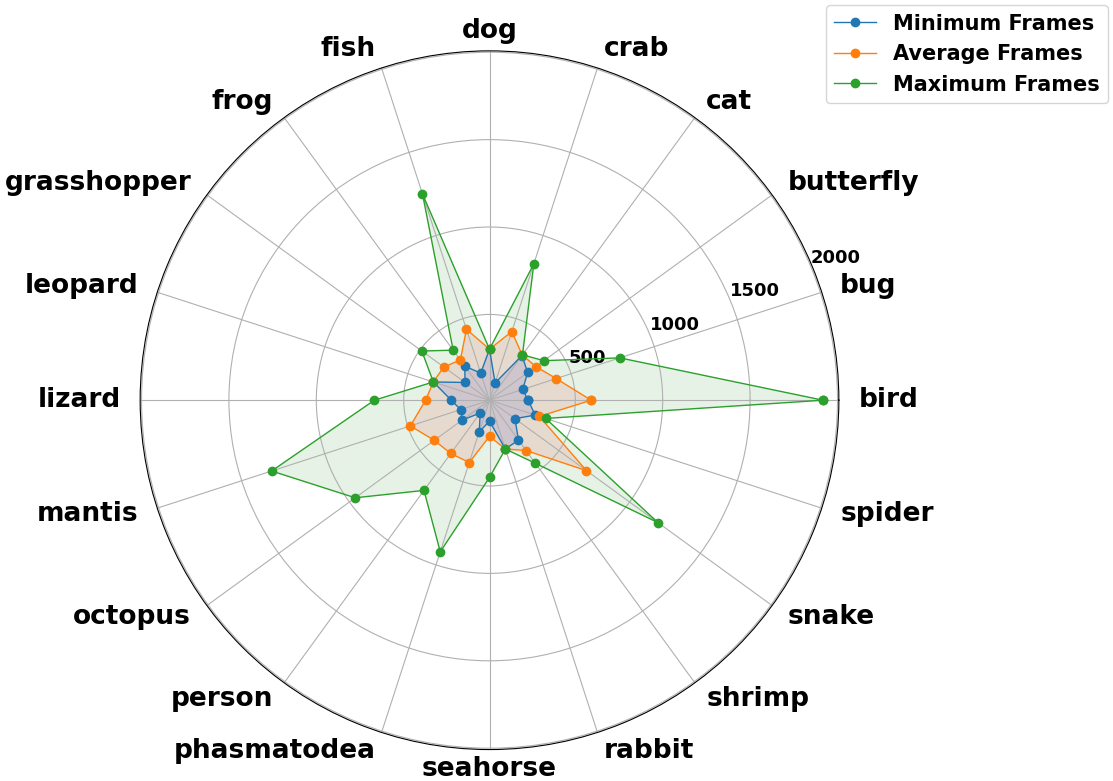}
   \caption{Radar chart compares min, avg, and max video frames in 20 categories. Blue, orange, green lines show distributions, highlighting differences.}
   \label{fig:oneco2}
   \vspace{-0.18in}
\end{figure}

In the video selection process, we apply stringent criteria to ensure the high quality and practicality of the data. Specifically, videos are evaluated and selected based on three main criteria: 1) the diversity and complexity of the environment, 2) the frequency with which targets cross frame boundaries, and 3) the overall length and continuity of the videos. We prioritize videos that present additional challenges, such as a high degree of similarity between targets and backgrounds. Due to resource constraints in collecting, processing, and annotating video data, we select 200 sequences for the benchmark test. This ensures that each video is of high quality with comprehensive annotations, adhering to our available time, budget, and manpower constraints.
The COTD dataset comprises 20 categories, namely: bird, bug, butterfly, cat, crab, dog, fish, frog, grasshopper, leopard, lizard, mantis, octopus, person, phasmatodea, seahorse, rabbit, shrimp, snake, and spider. To visually represent the composition of the dataset, we create a chart illustrating the number of sequences per category. Additionally, we generate a series of graphs depicting the maximum, minimum, and average number of frames for each category. These visual aids not only enhance the understanding of the dataset's diversity but also highlight the challenges in tracking camouflaged objects across different scenarios. 
Fig. \ref{fig:oneco2} provides a detailed view of the frame characteristics for each category, such as the range and average number of frames. These figures are crucial in demonstrating the dataset's comprehensiveness and its potential as a robust benchmark for evaluating the performance of camouflaged object tracking algorithms. Additionally, to highlight the challenges associated with tracking camouflaged objects, we evaluate seven state-of-the-art trackers on the COTD dataset and compare them with two popular generic object tracking benchmarks. The AUC results are presented in Table \ref{tab:3}. Notably, the AUC of these trackers on the TrackingNet benchmark is more than 15.0\% higher than on the COTD dataset, 
underscoring the significant challenges posed by tracking camouflaged objects.

\begin{table}[]
\centering
\caption{ AUC (\%) Comparison of state-of-the-art trackers on the COTD and generic object tracking benchmarks. }
\label{tab:3}
\vspace{-0.1in}
	\resizebox{3.3in}{0.6in}{
\begin{tabular}{@{}cccccccccc@{}}
\toprule
Dataset     & COTD     & TrackingNet \cite{muller2018trackingnet}     & LaSOT \cite{fan2019lasot}                \\\hline
HIPTrack \cite{HIPTrack}         & \textbf{68.0} & 84.5& 72.7    \\
 ROMTrack \cite{ROMTrack}         & \textbf{66.3} & 84.1 & 71.4      \\
DropTrack \cite{DropTrack}        & \textbf{67.4} &84.1& 71.8      \\
ARTrack \cite{ARTrack}         & \textbf{66.7} & 85.1 & 72.6      \\
SeqTrack \cite{SeqTrack}       & \textbf{64.1} & 83.9&71.5     \\
GRM \cite{GRM}   & \textbf{64.8}          & 84.0         & 69.9                      \\
SimTrack \cite{SimTrack}        & \textbf{63.9}         & 83.4         &  70.5             \\\bottomrule           
\end{tabular}
}\vspace{-0.22in}
\end{table}

\begin{table*}
\centering
\renewcommand{\arraystretch}{1.2} 
 \caption{Comparison of current popular benchmarks for visual single object tracking in the literature. $\ast$ indicates that only the test set of each dataset is reported.}
 \label{tab:benchmark}
\resizebox{1\textwidth}{!}{
\begin{tabular}{cccccccccc} 
\hline
Benchmark & Year & Videos & Total frames & classes & Attr.  & Avg.frames & Resolution  & Visual annotation                                                                                & Feature          
\\\hline
OTB100 \cite{Wu2015ObjectTB}  & 2015   & 100    & 59K          & 16             & 11 & 590    & 128$\times$96 $ \sim $ 800$\times$336  & BBox   & short-term        \\
UAV123 \cite{mueller2016benchmark}  & 2016   & 123    & 113K         & 9             & 12  & 915    & 720$\times$480 $ \sim $ 1280$\times$720 & BBox   & UAV  \\
DTB70 \cite{li2017visual} & 2017  & 70    & 15.8K          & -             & 11  & 225    & 1280$\times$720 $ \sim $ 1280$\times$720  & BBox  & UAV          \\
UAVDT \cite{du2018the}  & 2018   & 50    & 37.1K         & 3             & 8  & 742    & 960$\times$540 $ \sim $ 1024$\times$540 & BBox   & UAV  \\
WebUAV3M \cite{zhang2022webuav}  & 2022   & 4500    & 3.3M         &223            & 17 & 710    & 608$\times$320 $ \sim $ 1982$\times$1088 & BBox   & UAV  \\
TrackingNet* \cite{muller2018trackingnet} & 2018  & 511    & 226K          & 27             & 15  & 441.47   & 270$\times$360 $ \sim $ 1280$\times$720  & BBox  & large scale          \\
LaSOT* \cite{fan2019lasot}  & 2019   & 280    & 685K         & 85             & 14  & 2447.71    & 202$\times$360 $ \sim $ 1280$\times$720 & BBox   & category balance  \\
GOT-10k* \cite{huang2019got} & 2019  & 180    & 23K          & 84             & 6  & 133.33    & 270$\times$480 $ \sim $ 3840$\times$2160  & BBox  & generic          \\
VOT \cite{kristan2016novel}      & 2016  & 60     & 21K          & 37             & 9  & 357.58    & 320$\times$180 $ \sim $ 1280$\times$720   & BBox  & annual            \\
TSFMO250 \cite{zhang2022tracking}      & 2022  & 250     & 51K          & 26             & 12  & 202.72    & 720$\times$404 $ \sim $ 1310$\times$720   & BBox  & small and fast moving            \\
TRO \cite{Guo2024TrackingRO}      & 2024  & 200     & 70K          & -             & 12  & 349.03    & 244$\times$578 $ \sim $ 2400$\times$1080   & BBox  & reflected            \\
LLOL \cite{zhong2024low}      & 2024  & 269     & 133K          & 32            & 12  & 493.66    & 544$\times$960 $ \sim $ 3840$\times$2160   & BBox  & low-light            \\
 MoCA-Mask \cite{cheng2022implicit} & 2022      & 87     &  23K          & 45             & - & 263.67      & 480$\times$360 $ \sim $ 1920$\times$1080 & Mask   & camouflaged            \\
BihoT \cite{wang2024bihot}   & 2024     & 49     & 42K          & -             & 9  & 855.35    & - & BBox   & hyperspectral camouflaged            \\
UW-COT220 \cite{zhangunderwater}   & 2025     & 220     & 159K          & 96             & -  & 722    & 320$\times$240 $ \sim $ 3840$\times$2160 & BBox+Mask   & underwater camouflaged           \\
\textbf{COTD (Ours)}   & 2025    & 200    & 80K          & 20             & 12  &406.55   & 270$\times$480 $ \sim $ 2488$\times$1152  & BBox  & camouflaged                \\\bottomrule
\end{tabular}
}
\end{table*}

Compared to existing benchmark datasets, the COTD (ours) dataset focuses on tracking camouflaged objects, which presents a unique challenge in the field of visual single-object tracking. A summarized comparison with existing benchmarks is reported in Table \ref{tab:benchmark}. The COTD dataset includes 200 videos and 80,000 frames, surpassing in quantity the 100 videos and 59K frames of OTB100 \cite{Wu2015ObjectTB}, as well as the 70 videos and 15.8K frames of DTB70 \cite{li2017visual}. Furthermore, compared to BihOT \cite{wang2024bihot} and UW-COT220 \cite{zhangunderwater}, COTD has a higher focus on camouflage tracking. BihOT contains 49 videos and 42K frames, specializing in hyperspectral camouflage, while UW-COT220, although it includes 220 videos and 159K frames, focuses on underwater camouflage. However, the scale and diversity of the COTD dataset enable it to better simulate the complexity and challenges of tracking camouflaged objects in real-world scenarios. COTD specializes in tracking camouflaged objects in various environments, contrasting with the application scenarios of BihOT and UW-COT220.

\subsection{Data Annotation}
We follow the principles in \cite{muller2018trackingnet} for sequence annotation. Each video frame is meticulously annotated by a team of expert annotators, specifically students engage in visual tracking research. Annotators draw and edit axis-aligned bounding boxes on each frame to represent the most compact boundary of the object. Visible targets are annotated as present, while those missing, out of view, or occluded are marked accordingly. To enhance the efficiency and quality of the annotations, we adopt a stringent manual annotation process. Each frame's annotation underwent thorough review to ensure the accuracy and consistency of the bounding boxes. Annotators meticulously label each object frame-by-frame, ensuring accurate identification even within complex backgrounds. To further ensure high-quality annotations, we implement a double-check mechanism where a second annotator review the initial annotations to resolve any discrepancies. All differences are addressed through discussion and consensus. 
This rigorous manual annotation process and double-check mechanism guarantee the high quality of the dataset.  
Examples of bounding box annotations in the COTD are shown in Fig. \ref{fig:Anno}.

\begin{figure}
  \centering
\includegraphics[width=0.46\textwidth,height=0.24\textwidth]{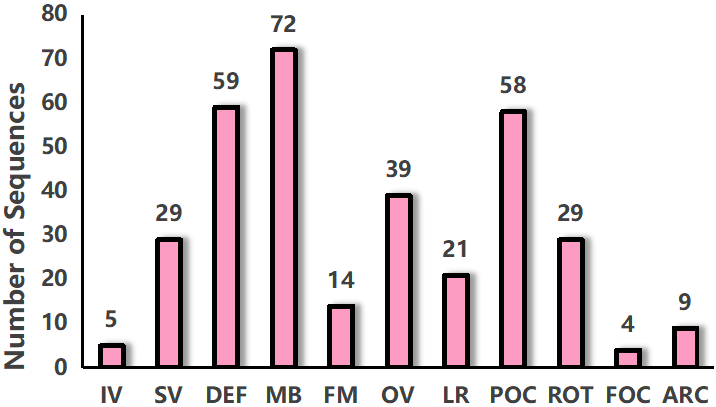}
  \caption{Distribution of sequences with each attribute in COTD.}
  \vspace{-0.15in}
  \label{fig:Attribute-compar}
\end{figure}

\subsection{Attributes}
\begin{figure*}[t]
\centering
\includegraphics[width=0.98\textwidth]{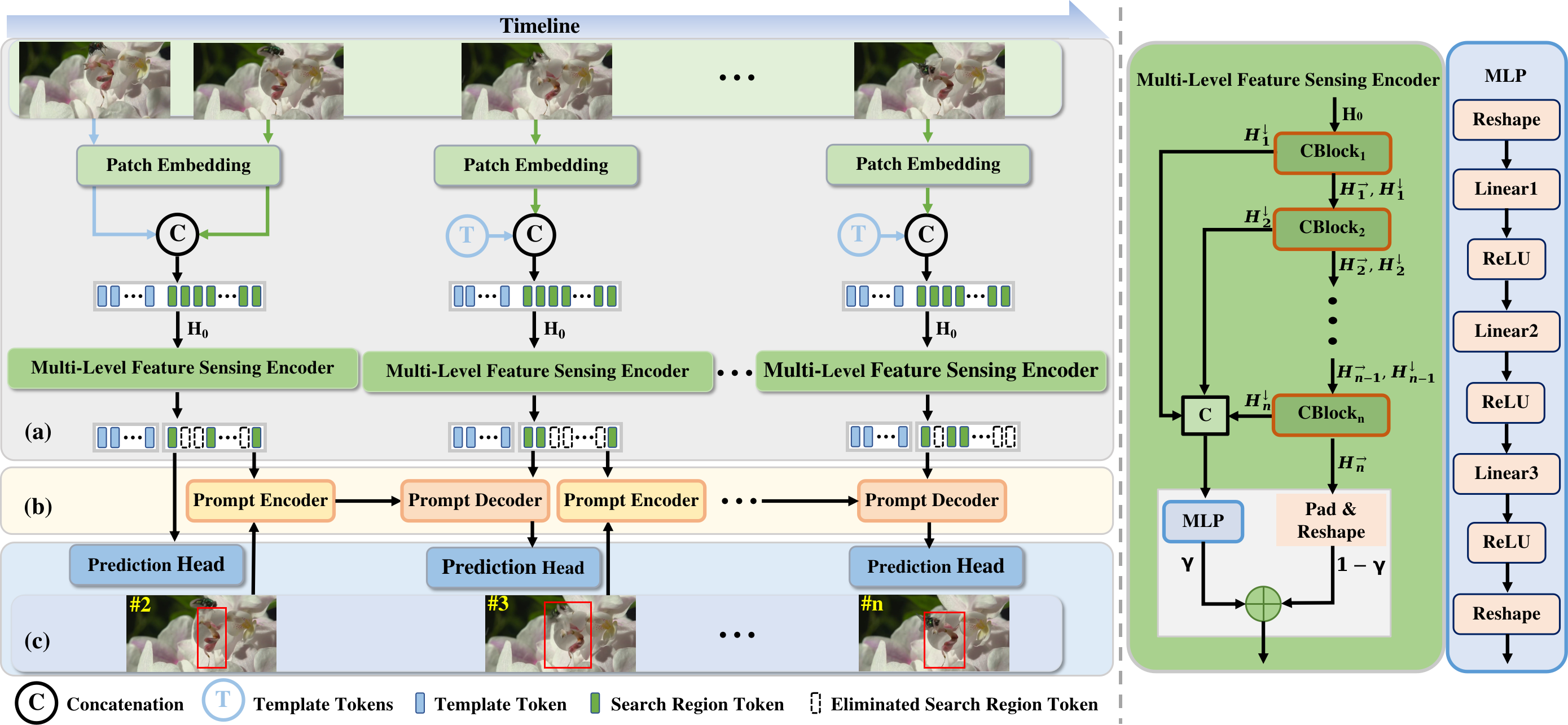} \vspace{-0.15in}
\caption{The framework of HIPTrack-MLS, which inherits from HIPTrack but uses a Multi-Level Feature Sensing Encoder in the feature extraction network. The detailed structure of this encoder is shown on the right.}
\vspace{-0.1in}
\label{fig:hip-MLS}

\end{figure*}

Following the prevalent tracking benchmark methodologies \cite{zhu2021detection}, \cite{huang2019got}, and \cite{wu2015object}, we meticulously annotate each video sequence with detailed attributes to enable an in-depth analysis of tracking algorithms' performance. The proposed COTD dataset features 12 attributes, with each sequence including Background Clutter (BC) to highlight the dataset's camouflage characteristics. In addition, we annotate the following 11 attributes: (1) Illumination Variation (IV), (2) Scale Variation (SV), (3) Deformation (DEF), (4) Motion Blur (MB), (5) Fast Motion (FM), which is marked when the object center moves by at least 20\% of its own size between consecutive frames, (6) Out-of-View (OV), (7) Low Resolution (LR), marked when the object area is smaller than 900 pixels, (8) Partial Occlusion (POC), (9) Rotation (ROT), (10) Full Occlusion (FOC), and (11) Aspect Ratio Change (ARC), marked when the object's bounding box aspect ratio falls outside the range [0.5, 2]. Fig.\ref{fig:Attribute-compar} shows the distribution of these attributes on COTD.  As can be seen, the most common challenge in COTD is Motion Blur. In addition,
the Deformation and Partial Occlusion also present frequently in COTD. By comprehensively considering these attributes, we can conduct a thorough evaluation of tracking algorithms under various challenging conditions.

 \section{Method}

In our evaluation, HIPTrack \cite{HIPTrack} outperforms other advanced trackers in COTD but still falls short of satisfactory levels. Camouflaged objects' similar textures, colors, and patterns to their environment create visual ambiguity, making single-level feature representation insufficient for detection. To address this, we propose HIPTrack-MLS, a novel baseline tracker based on HIPTrack that enables multi-level feature sensing. By capturing information at various scales and resolutions, multi-level features enhance HIPTrack's ability to distinguish targets from their environment, significantly improving its representation power and increasing the accuracy of identifying camouflaged objects.

\subsection{Overall Architecture}

Similar to HIPTrack \cite{HIPTrack}, HIPTrack-MLS comprises three main components, as illustrated in Fig. \ref{fig:hip-MLS}, left panel, and labeled as (a), (b), and (c), respectively.
The feature extraction network extracts features from the search region, filtering out background patches. The historical prompt network, with an encoder and decoder, encodes and stores target features and generates prompts combined with search region features. The head network is the same as OSTrack's. The key difference is in the feature extraction network: HIPTrack-MLS uses a novel Multi-Level Feature Sensing Encoder ( Fig. \ref{fig:hip-MLS}, right panel) to enhance feature representation for camouflaged objects by perceiving multi-level features. Other inherited architectures are detailed in \cite{HIPTrack}.
\subsection{Multi-Level Feature Sensing Encoder}

Our Multi-Level Feature Sensing Encoder extends the Transformer Encoder architecture in HIPTrack, specifically being a Vision Transformer (ViT) with early candidate elimination modules. It consists of stacked CEBlocks, as shown in Fig. \ref{fig:hip-MLS}, right panel. The key innovation is the Multi-Level Feature Sensing structure that processes multi-level features from ViT blocks. In this design, features from all levels are concatenated, enabling the model to leverage both detailed low-level and abstract high-level information. A multilayer perceptron with three linear layers and ReLU activations reduces the dimensionality from 12 to 6, then to 3, and finally to 1, enhancing the model's ability to capture complex patterns. To handle potential shape changes due to early candidate elimination, CEBlock is split into two parts: $\mathbf{CBlock}^{\rightarrow}_i$ with elimination and $\mathbf{CBlock}^{\downarrow}_i$ without it, with corresponding outputs $H^{\rightarrow}_i$ and $H^{\downarrow}_i$. The effect of our proposed Multi-Level Feature Sensing Encoder can be expressed as:
\begin{equation}
{F}_{mls}=(1-\gamma)\mathfrak{P}\left(H_{n}^{\rightarrow}\right)+\gamma \text{MLP}\left([H^{\downarrow}_{1},..., H^{\downarrow}_{n}]\right),
\end{equation}
where $H^{\downarrow}_{i}=\mathbf{CBlock}^{\downarrow}_i(H^{\downarrow}_{i-1})$, $H^{\rightarrow }_{i}=\mathbf{CBlock}^{\rightarrow }_i(H^{\rightarrow }_{i-1})$, $\mathfrak{P}$ represents padding and reshaping used to align $H^{\rightarrow }_n$ with $H_0$, and $\gamma \in [0,1]$ is a weighting constant balancing the importance of the aggregated feature and $H^{\rightarrow }_n$. $\mathbf{MLP}$ denotes the multilayer perceptron network.
When $\gamma=0$, $F_{mls}$ is simply the feature produced by the Transformer Encoder of HIPTrack. Thus, our Multi-Level Feature Sensing Encoder extends HIPTrack’s feature extraction network, enhancing feature representation by capturing information at multiple levels of abstraction. Since the proposed encoder does not introduce additional training loss, we can use the same training pipeline as HIPTrack for training HIPTrack-MLS. For details, please refer to HIPTrack \cite{HIPTrack}.

\section{Evaluation}
 
We use one-pass evaluation (OPE) to measure precision and success rate, following \cite{Fan2019LaSOTAH,Mller2018TrackingNetAL}. Precision is the pixel distance between estimated and ground truth bounding box centers, and success rate is based on IoU exceeding 0.5 \cite{Wu2015ObjectTB,Fan2019LaSOTAH,Mller2018TrackingNetAL}. Trackers are ranked by precision at 20 pixels (PRC) and success plot's AUC.

			


\subsection{Evaluation Results}

\begin{figure}[h]
	\centering
	{
		\begin{minipage}[t]{0.23\textwidth}
			\includegraphics[width=1\textwidth]{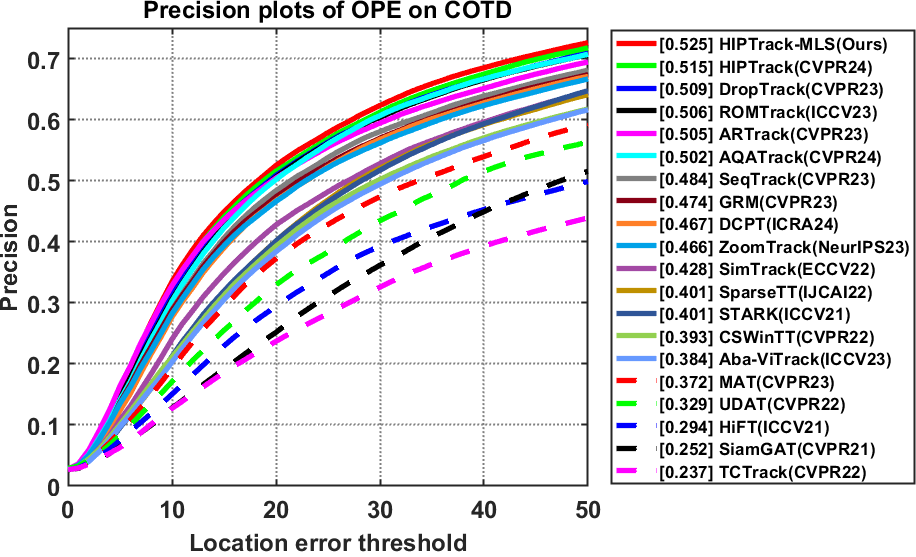}\hspace{0in}
			
	\end{minipage}}
	{
		\begin{minipage}[t]{0.23\textwidth}
			\includegraphics[width=1\textwidth]{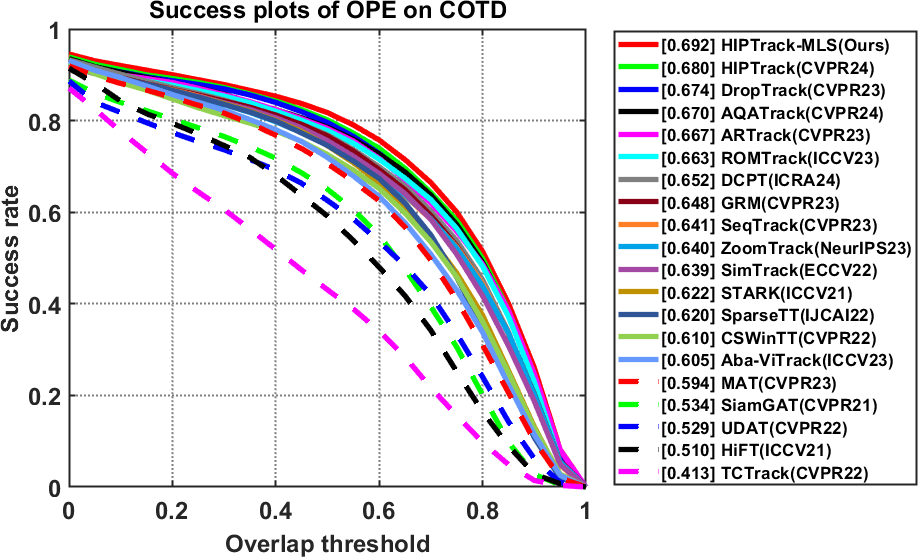}\hspace{0in}		
	\end{minipage}}
 \vspace{-0.15in}
	\caption{Overall performance on COTD. Precision and success rate for one-pass evaluation (OPE) \cite{wu2013online} are used for evaluation. }
	\label{fig:result}\vspace{-0.15in}
\end{figure}

\noindent\textbf{Overall Performance:}We evaluated 20 state-of-the-art trackers and our HIPTrack-MLS using the COTD dataset, with results shown in Fig. \ref{fig:result}. HIPTrack-MLS outperforms all other trackers, achieving a precision (PRC) of 0.525, 1.0\% higher than its baseline HIPTrack (0.515), and an AUC of 0.692, surpassing HIPTrack's 0.680 by 1.2\%. This highlights HIPTrack-MLS's superior performance in both metrics, thanks to its advanced multi-level feature sensing.

			
			
 

\begin{figure}[h]
	\centering
	{
		\begin{minipage}[t]{0.23\textwidth}
			\includegraphics[width=1\textwidth]{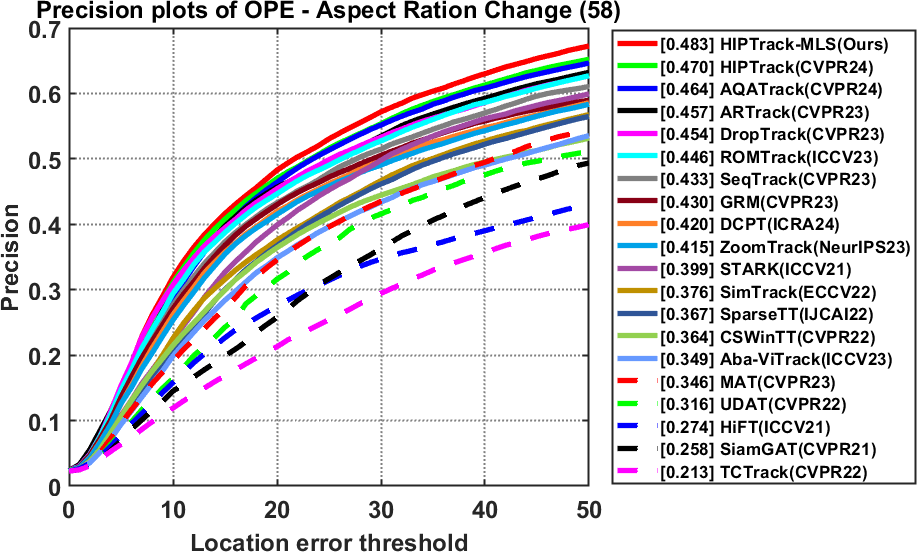}\hspace{0in}		
	\end{minipage}}
 {
		\begin{minipage}[t]{0.23\textwidth}
			\includegraphics[width=1\textwidth]{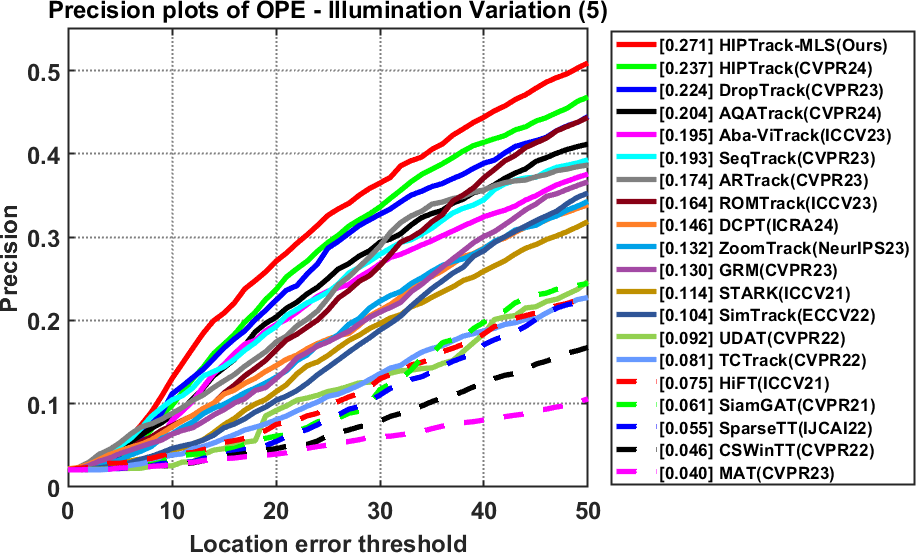}\hspace{0in}
	\end{minipage}}
 {
		\begin{minipage}[t]{0.23\textwidth}
			\includegraphics[width=1\textwidth]{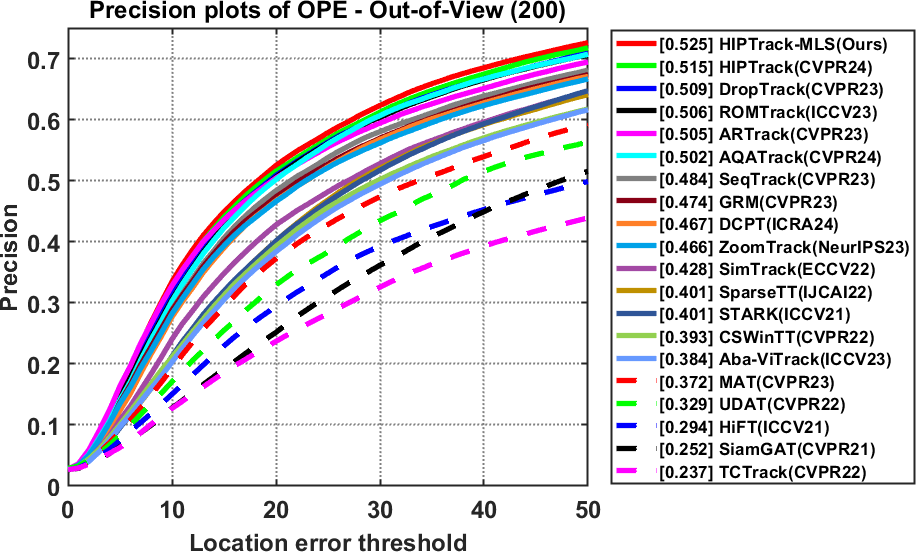}\hspace{0in}
			
	\end{minipage}}
 {
		\begin{minipage}[t]{0.23\textwidth}
			\includegraphics[width=1\textwidth]{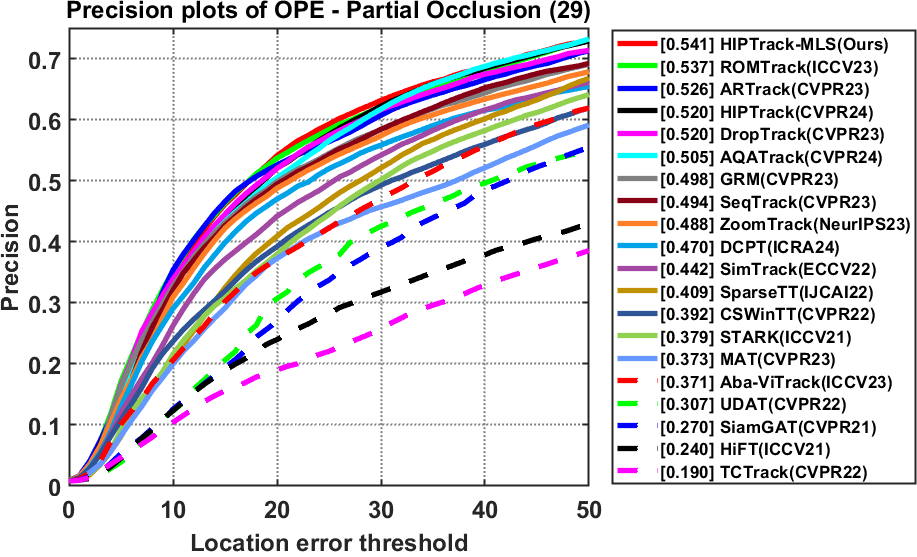}\hspace{0in}
			
	\end{minipage}}
 
 \vspace{-0.15in}
	\caption{Attribute-based comparison on Aspect Ration Change, Illumination Variation, Out-of-View, and Partial Occlusion. }
	\label{fig:Attribute}\vspace{-0.15in}
\end{figure}

\noindent\textbf{Attribute-based performance:} We evaluated the performance of different trackers on the COTD dataset across eleven common attributes. HIPTrack-MLS consistently ranks first in most attribute subsets. Due to space constraints, we present PRC results for four challenges in Fig. \ref{fig:Attribute}: Aspect Ratio Change (ARC), Illumination Variation (IV), Out-of-View (OV), and Partial Occlusion (POC). More results are in the supplementary material. HIPTrack-MLS ranks first in all these subsets, surpassing HIPTrack by at least 1.0\%. Notably, it achieves a precision of 0.271 in IV, beating HIPTrack by 3.4\%, and shows improvements of 1.3\% in ARC and 1.0\% in OV. These results highlight the effectiveness of our method for tracking camouflaged objects.

\begin{figure}[h]
	\centering
	\includegraphics[width=0.475\textwidth]{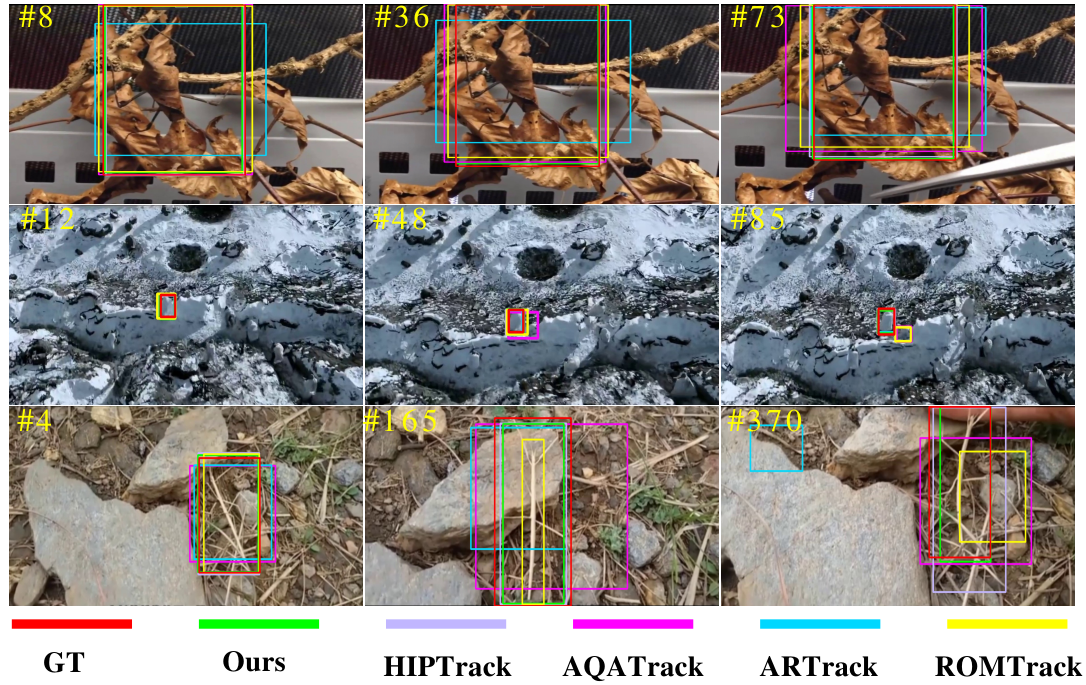}	
	\caption{Qualitative evaluation on 3 COTD sequences: mantis-2, person-9, and phasmatodea-10. Different colors show results of various methods, with `GT’ for ground truth.}
	\label{fig:Qualitative evaluation}
\end{figure}


\noindent\textbf{Qualitative Evaluation:} In Fig. \ref{fig:Qualitative evaluation}, we compare our method with 4 top trackers facing challenges like Illumination Variation, Deformation, Rotation, Aspect Ratio Change, and Partial Occlusion. In mantis-2, HIPTrack-MLS shows superior performance. In person-9, only HIPTrack-MLS, HIPTrack, and  ARTrack can track the target amidst clutter and interference. In phasmatodea-10, many trackers fail due to background effects, while HIPTrack-MLS remains accurate. These results highlight our method's effectiveness for tracking camouflaged objects.

\subsection{Ablation Study}
\noindent\textbf{Impact of the importance coefficient in Multi-Level Feature Sensing}. We evaluated HIPTrack-MLS on the COTD dataset with $\gamma$ ranging from 0.1 to 1.0 in increments of 0.1 to study its effect on balancing aggregated multi-level features and $H_{n}^{\rightarrow}$. As $\gamma$ increases, the aggregated feature's influence grows while reliance on $H_{n}^{\rightarrow}$ diminishes. Table \ref{tab:4} shows PRC and AUC results for different $\gamma$ values. The optimal PRC (52.5\%) and AUC (69.2\%) are achieved at $\gamma=0.1$, indicating that an appropriate $\gamma$ choice enhances tracking performance. Thus, we set $\gamma=0.1$ as the default for HIPTrack-MLS.

\begin{table}[t]
\centering
\caption{Shows the effect of the aggregated feature's importance coefficient on HIPTrack-MLS's PRC (\%) and AUC (\%) on COTD. Red, blue, and green denote 1st, 2nd, and 3rd place.}
\label{tab:4}
\vspace{-0.1in}
\resizebox{3in}{0.25in}{%
\begin{tabular}{@{}cccccccccccc@{}}
\toprule
$\gamma$ & 1.0 & 0.9 & 0.8 & 0.7 & 0.6 & 0.5 & 0.4 & 0.3 & 0.2 & 0.1 \\ \midrule
PRC (\%) & 47.0 & 48.1 & 49.1 & 49.9 & 50.4 & 51.1 & 51.7 & {\color{green} \textbf{52.1}} & {\color{blue} \textbf{52.2}} & {\color{red} \textbf{52.5}} \\
AUC (\%) & 66.7 & 67.4 & 68.0 & {\color{blue} \textbf{69.1}} & 68.1 & 68.2 & 68.7 & {\color{green} \textbf{68.8}} & 68.3 & {\color{red} \textbf{69.2}} \\ \bottomrule
\end{tabular}%
}
\vspace{-0.12in}
\end{table}

\section{Conclusion}

This work explores the underexplored task of tracking camouflaged objects, which has significant application potential. Currently, no public benchmark exists for this task. We introduce COTD, the first benchmark for camouflaged object tracking. We also evaluated 20 state-of-the-art tracking algorithms and introduced HIPTrack-MLS, an improved version of HIPTrack with enhanced multi-level feature sensing. Experiments show HIPTrack-MLS outperforms leading algorithms on COTD.

\bibliographystyle{IEEEtran}
\bibliography{main.bib}

\end{document}